\documentclass[runningheads]{llncs}
\usepackage{graphicx}
\usepackage{comment}
\usepackage{amsmath,amssymb} 
\usepackage{color}
\usepackage{float}
\usepackage{subcaption}
\usepackage{tabu,multirow}
\usepackage[width=122mm,left=12mm,paperwidth=146mm,height=193mm,top=12mm,paperheight=217mm]{geometry}

\begin{document}
\pagestyle{headings}
\mainmatter
\def\ECCVSubNumber{7433}  

\title{Applying Tensor Decomposition for the Robustness against Adversarial Attack} 


\author{Seungju Cho, Tae Joon Jun, Mingu Kang, Daeyoung Kim}

\maketitle
\vspace{2cm}
\begin{abstract}
Nowadays, deep learning technology is growing faster and shows dramatic performance in computer vision. However, it turns out that the deep learning model is highly vulnerable to small perturbation called an \textit{adversarial attack}. 
So far, although many of the \textit{defense mechanism} has been proposed to mitigate the effect of the adversarial attack, all of them are under rigorous assumptions. 
However, our approach is not tied up any assumptions since our insight stems from the \textit{Tensor Decomposition}.
In this paper, we experimentally demonstrated that decomposing the tensor would be an effective countermeasure against several adversarial attacks. We conducted experiments with well-known benchmarks such as MNIST, CIFAR-10, and ImageNet dataset. 
Our experimental results show that this simple method has capable of having attack resilience and robustness against adversarial attacks.
To the best of our knowledge, this is the first approach to leverage the tensor decomposition as a defense mechanism.
We hope that leveraging the tensor decomposition becomes a universal approach to solve inherent corner cases of deep learning models. 
\end{abstract}

\section{Introduction}

Over the past several years, advances in deep neural networks (DNNs) have widely expanded the ability of what the machine can deal with.
Especially, DNNs have achieved remarkable successes for image classification  \cite{krizhevsky2012imagenet,sermanet2013overfeat} and it even goes beyond human capability \cite{he2016deep}. 
With this performance, deep learning technology has started to be applied to various areas.
However, some papers \cite{szegedy2013intriguing,goodfellow2014explaining,carlini2017towards,kurakin2016adversarial,kurakin2016adversarial_scale,moosavi2016deepfool,Eykholt_2018_CVPR,Chen2017EADEA} proved that even DNNs can be easily fooled by small changes to input that is imperceptible to a human eye. 
According to these studies, carefully crafted perturbations to the vision-based applications can induce systems to behave in unexpected ways.
Indeed, this is small enough to be inconspicuous, but some researches show that its influence might be more than expected since even state-of-the-art models get an almost zero-classification accuracy under \cite{carlini2017towards}.

\begin{figure}[H]
\centering
\includegraphics[width = 10cm]{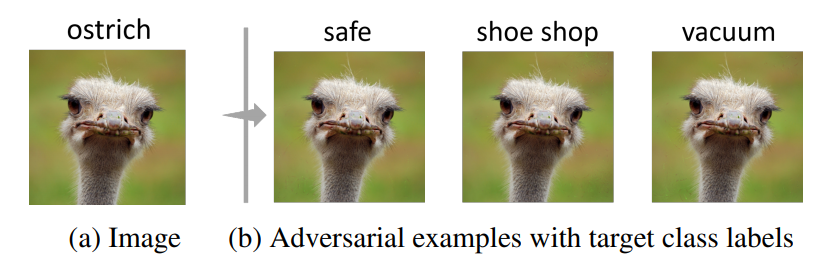}
\caption{Adversarial Examples crafted by an attack algorithm \cite{Chen2017EADEA}}
 \label{fig:AdvExample}
\end{figure}

 Considering the deep learning models do not hesitate whenever judge the output, it might cause crucial accidents.
For instance, Fig.~\ref{fig:AdvExample} represents the adversarial examples in the image classification task. 
Although all of the images can be seen as an ostrich by human visible intuition, deep learning model outputs clearly different labels due to lack of such intuition.  
From a more theoretical perspective, misclassification occurs when the adversarial perturbations cross the decision boundary, but the existing classifier has no such intuition that can ward it off.
Now, the corner case of DNNs which have been alluded to adversarial attacks is getting pervasive and being more sophisticated.
As a result, the vulnerability of adversarial attacks hinders its adoption for some safety-critical system and also security-sensitive application, including an autonomous-driving car \cite{Eykholt_2018_CVPR,sitawarin2018darts}.
Since the advent of such adversarial attacks, many researchers or vendors have paid significant attention to adversarial examples.
This is because they might not want to go through all the risks of their applications or models. They might as well choose to verify the robustness rather than take risks.
However, the resistance against adversarial examples renders another challenge, for no method can be a cure-all against adversarial attacks.
To make up for corner cases of DNNs, several studies \cite{goodfellow2014explaining,szegedy2013intriguing,meng2017magnet,papernot2016distillation,song2017pixeldefend,jia2019comdefend,Liao_2018_CVPR,xie2019feature} have proposed the defense mechanism against adversarial attacks to mitigate the potential of the risk by adversary.
These defense mechanisms can be viewed as two main approaches:
(1) changing the model itself, which can improve the robustness by training with adversarial examples, e.g., \textit{adversarial training} \cite{goodfellow2014explaining,szegedy2013intriguing,tramer2017ensemble}, (2) preprocessing the inputs to diminish the effect of adversarial noise, 
e.g., \textit{Magnet, Comdefend, PixelDefend, Defense-GAN, HGD, etc.}  \cite{meng2017magnet,jia2019comdefend,song2017pixeldefend,samangouei2018defense,Liao_2018_CVPR,xie2019feature}.

However, (1) are designed to deal with specific adversarial attack strategies in mind, so generalization is likely to be restricted. 
It implies that the models using this method may be vulnerable to another attack optimized with such attack strategies.
On the other hand, (2) utilize models with a vast amount of legitimate data to purify the inputs itself, instead of assuming some attack strategies.
The main point of (2) is to measure the distance between the inputs and manifold of the legitimate images, and then approximate or guide the adversarial images closer to the manifold of the legitimate images.
To purify the inputs, therefore, a well-generalized model should be required to assure the performance. 
Given that the adversarial images occupy the low probability region trained with legitimate samples \cite{song2017pixeldefend}, poorly generalized models cannot ensure that the aforementioned approaches get a good result. 
Also, another attack technique might be created by an adversary who knows the model's structure.
In a nutshell, as these approaches are likely to be a temporary expedient, the universal defense approaches which can cover a myriad of risk should be explored.

Here, we propose a novel intuition for deep learning models, which can make the model universally robust.
To the best of our knowledge, this is the first approach to explore the defense in terms of the universal point of view.
Our approach leverages the potential power of the tensor decomposition to diminish the effect of adversarial noise by using the reconstructed images as an input of a deep learning model.
The reconstructed inputs are fed into the classifier, and we experimentally demonstrate that such simple preprocessing could be an effective countermeasure against the adversarial attack. On MNIST, a degradation of top-1 accuracy on adversarial example is less than 1\% against four adversarial attacks and less than 10 \%  on CIFAR-10 and ImageNet. This result outperforms recent defense mechanisms \cite{Liao_2018_CVPR,jia2019comdefend}.
To ensure that deep learning applications extend their potential of utilization toward other domains, it would be better to take into account the robustness of those applications.
If you want to avoid cherry-picking doubt and make the model more general across a variety of risks including adversarial attacks, our insight would be an interesting candidate. 
Our contribution is as follows:
\begin{enumerate}
  \item \textbf{High Compatibility. } Our approach leverages the tensor decomposition for preprocessing the inputs which might have been affected by the adversary. We do not assume anything such as attack strategies or classifiers, just use an input as a reconstructed input by using the tensor decomposition method, which indicates that our proposed method can be relatively easy to utilize and be applied to whatever the classifier is. 
  \item \textbf{Efficient Engineering Complexity. } As we mentioned above, tensor decomposition just depends on what the input it is, so it is not tied up with the attack strategies or classifiers. Therefore, we do not need to focus on how the model classifies the input since tensor decomposition is free from the model dependency. It implies that retraining the model or augmenting the training data could no longer be required. It requires only processing time to reconstruct inputs. Even more, the processing time is negligible.
  \item \textbf{Integrity of the inputs. } When it comes to the reconstruction process, some information that in charge of the important role might be lost. Although state-of-the-art approaches \cite{Liao_2018_CVPR,jia2019comdefend} have gotten remarkable performance, their proposed model degrades the performance with even the clean images. This is some kind of a trade-off. It thus makes it difficult to apply defense mechanisms. However, tensor decomposition could incur less adverse effects on the clean images, and ensure its performance even at the high-dimension dataset such as ImageNet.
\end{enumerate}

\newpage
\section{Related work}
\paragraph{\textbf{Adversarial Attacks. }}Szegedy \textit{et al.} found the existence of adversarial perturbation that
breaks the image classifier thorough solving adversarial optimization problem \cite{szegedy2013intriguing,goodfellow2014explaining}. They show the model accuracy is dropped even though the perturbed image looks similar to human eyes. 
Goodfellow \textit{et al.} \cite{goodfellow2014explaining} uses the sign of the gradient of input with respect to the loss function of the target model. 
This method is called Fast Gradient Sign Method (FGSM) since it updates input once. 
With a similar idea, \cite{kurakin2016adversarial} uses FGSM in an iterative way. 
Chen \textit{et al.} \cite{Chen2017EADEA} leverages $L_{1}$ distortion to generate effective adversarial examples and improve the attack transferability which refers to the attack success rate using the adversarial examples which come from the substitute models. In other words, high transferability implies that the performance of the target model might depreciate even without the knowledge about the model, i.e., \textit{black box attack}.
Carlini and Wagner \cite{carlini2017towards} changes the optimization problem defined in \cite{szegedy2013intriguing} for achieving more powerful attack. 
\cite{moosavi2016deepfool} measures the minimum size required for the attack. 
They approximate the decision boundary of the model and update input repeatably until the model misclassifies it. 
Besides the image classification task, \cite{Eykholt_2018_CVPR,sitawarin2018darts} demonstrated that adversarial attacks can be applied beyond the digital space, so security concerns could arise in even physical space such as the autonomous-driving car.

\paragraph{\textbf{Adversarial Defense.}} To counter adversarial attacks, some works trained the model with adversarial examples to ensure that the model has a resilience against those adversarial examples, which have been called \textit{adversarial training}. 
During the process of training, they generate adversarial images for improving the performance. 
Although it works, it depends on the particular adversarial data used in the training process. 
For instance, \cite{kurakin2016adversarial_scale} shows their approach is robust in the simple attack, but not in a more sophisticated attack. In addition, it has an engineering penalty since it requires retraining the model. 
If it takes longer to create an adversarial example, it will take more time to retrain the model.
Instead of using the data augmentation, methods to change the model itself were also proposed \cite{papernot2016distillation}. 
They change the objective function of the problem for obtaining the robustness. 
However, this approach also has to retrain the model, so it also boils down to increasing the engineering complexity.

In recent years, several papers \cite{meng2017magnet,jia2019comdefend,song2017pixeldefend,Liao_2018_CVPR,xie2019feature,samangouei2018defense} preprocess the inputs before putting into the classifier. 
They propose the model which serves the direction to approximate the distribution of the adversarial images as close as possible to the decision boundary.
All of the methods require a well-generalized defense model, so the even clean images could be affected when the model is poorly generalized. 
It results in damage to the integrity of the inputs.
To guarantee the integrity of the model, all of them require well-generalized classifiers to detect if the input is adversarial or approximate the manifold of legitimate samples.
Our approach is similar to those approaches in terms of the preprocessing, yet differentiation is our method does not need any premises, including the detector or well-generalized models.
Consequently, our method does not hurt the performance in terms of integrity.


\newpage
\section{Background}

\subsection{Adversarial Attack}
Basically, all of the attacks use the gradient of data with respect to the loss function of the target model.
 In this section, we briefly review the basic method of adversarial attack.  
 
 \noindent \textbf{Fast Gradient Sign Method (FGSM):} The \textit{FGSM} is proposed by \cite{goodfellow2014explaining}. It is a simple and effective attack method. The image $X$ is perturbed as follows.
 \begin{equation}
     X = X + \epsilon \cdot  | sign(\triangledown_{X} l(X,y_{true})) |
 \end{equation}
Where $\epsilon$ is a magnitude of noise and $l(X,y_{true})$ is a loss with respect to the true label of the image.
It adjusts input X by adding a sign of the gradient of X. It increases the loss function of the target model so that the model misjudges the adjusted input. Since it updates input X once, it is also called \textit{single-step} method. 

 \noindent \textbf{Basic Iterative Method (BIM):}
The \textit{BIM} is a repetitive version of FGSM \cite{kurakin2016adversarial}. it is a more powerful attack method compared to the FGSM. And it is also called \textit{Iterative FGSM}. It uses the following equations:
\begin{equation}
   X_{t+1} = \text{clip}_{X,\epsilon}(X_{t} + \alpha  \cdot   | sign(\triangledown_{X} l(X_{t},y_{true})) |)
\end{equation}
Where   $X_{0} = X$, $\alpha$ is a step size for adjusting $X_t$ , and clip function ensures that $X_t \in (X-\epsilon,X+\epsilon )$ for all $t$.  It is also called \textit{multi-step} method. Here $\alpha$ is the $1$ in the scale of 0 to 255 in the original paper. 

\noindent \textbf{DeepFool:} DeepFool attack approximates the decision boundary of a classifier, and measure the minimal perturbations that are sufficient to fool the classifier \cite{moosavi2016deepfool}. For the affine multiclass classifier, they calculate the distance $d$ as follows. 
\begin{equation}
d = \frac{|f_k(x) - f_l(x)|}{w'} ~ \text{where}~  w' = \|\nabla f_k(x) - \nabla f_l(x) \|_2
\end{equation}
Here $f_k$ and $f_l$ are classifier for $k$ and $l$-th class. Similar to BIM, they update $x$ in an iterative way. For nonlinear classifiers, they approximate the linear boundary and find the distance to fool the nonlinear classifier.

\noindent \textbf{Carlini \& Wagner (C\&W):} Carlini and Wagner \cite{carlini2017towards} define an optimizaion problem to find an adversarial example. They define following optimizaion formulation. 
\begin{equation}
\text{min}_{\delta} ~ D(x,x+\delta) + c \cdot f(x+\delta) 
\end{equation}

\begin{figure}[H]
    \centering
  \begin{subfigure}{0.19\columnwidth}
  \includegraphics[width=\textwidth]{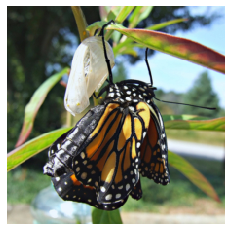}
  \caption{clean}
  \end{subfigure}
  \hfill
  \begin{subfigure}{0.19\columnwidth}
  \includegraphics[width=\textwidth]{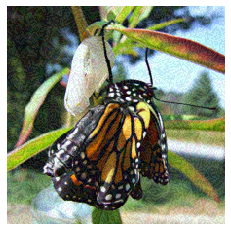}
  \caption{FGSM} 
  \end{subfigure} 
  \begin{subfigure}{0.19\columnwidth} 
  \includegraphics[width=\textwidth]{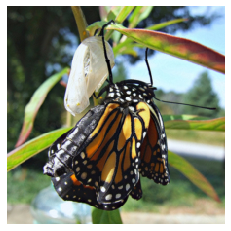}
  \caption{BIM} 
  \end{subfigure}  
  \hfill 
    \begin{subfigure}{0.19\columnwidth} 
  \includegraphics[width=\textwidth]{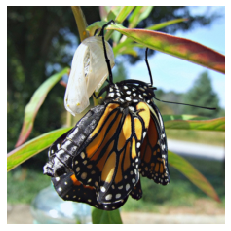}
  \caption{DeepFool} 
  \end{subfigure}  
  \hfill 
  \begin{subfigure}{0.19\columnwidth} 
  \includegraphics[width=\textwidth]{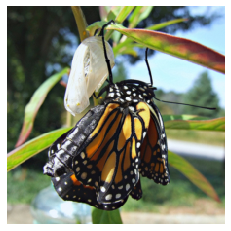}
  \caption{C\&W} 
  \end{subfigure}
  \caption{Visualization of adversarial examples on each method. The pre-trained Resnet101 \cite{he2016deep} model classifies clean image as a \textit{monarch}, image with FGSM as a \textit{sea slug}, image with BIM as a \textit{Doberman}, image with DeepFool as a \textit{hornbill} and image with C\&W as a \textit{longicorn}. However, we can check that theses images seem to the same in the human eye}
  \label{fig:advs}
  \end{figure}

Here $D$ is a distance metric to measure the distance between the clean image and adversarial image. $f(\cdot)$ is an objective function to control the result of original classifier $C$. C\&W attack is one of the most powerful attacks in this literature. We visualize the adversarial image generated by each method in Fig. \ref{fig:advs}
 
 \subsection{Tensor Decomposition}

\begin{figure}[H]
\centering
\includegraphics[width = 10cm,height = 3cm]{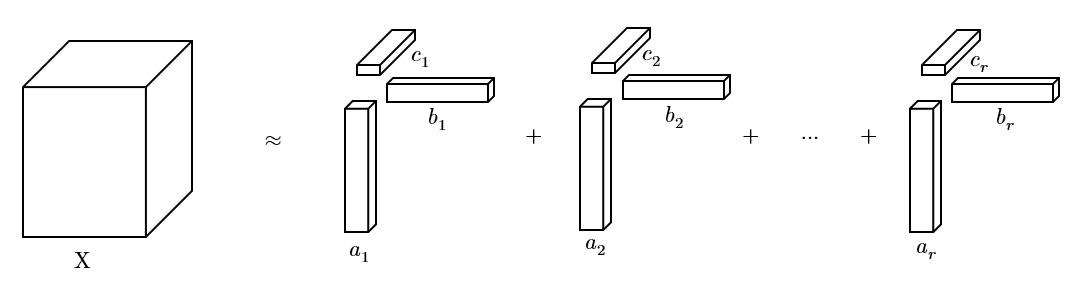}
\caption{\textit{CP} decomposition of tensor $X \in \mathbb{R}^{I \times J \times K}$. $X$ is approximately represented by the sum of the outer product of the tensor}
\label{fig:cp}
\end{figure}

 A \textit{tensor} is a multi-dimensional array. For instance, the color image is a tensor consists of height, width, and the color channel. A tensor decomposition method decomposes a tensor into low dimension tensors. The \textit{CANDECOMP/PARAFC} \cite{carroll1970analysis,harshman1970foundations} decomposition approximates a tensor $X$ as a sum of the outer product of the tensor belonging to each dimension as Fig.~\ref{fig:cp}. We refer to this as a \textit{CP} decomposition. 
\noindent Let $X \in \mathbb{R}^{I \times J \times K}$, $a_i \in \mathbb{R}^{I}, b_i \in \mathbb{R}^{J} \ \text{and}\ c_i \in \mathbb{R}^{K}$ for $i = 1,\dots,r$. Here $r$ is the number of components and it is a hyperparameter. If $r$ is small, tensor$X$ is approximated into low dimension tensor. So we call deciding $r$ as choosing the dimension for convenience. Then $x_{ijk}$ is approximated as follows. 
 \begin{equation}
x_{ijk} \approx \sum_{l = 1}^{r} a_{il}b_{jl}c_{kl}
 \end{equation}

\begin{figure}
\centering
\includegraphics[width = 7cm,height = 3cm]{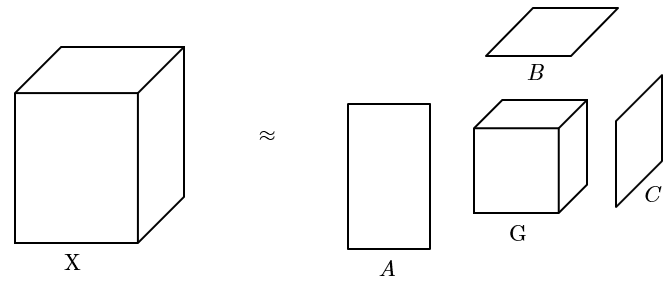}
\caption{\textit{Tucker} decomposition of tensor $X \in \mathbb{R}^{I \times J \times K}$. $X$ decomposed as a core tensor $G$ and factor tensors $A,B$ and $C$}
\label{fig:tucker}
\end{figure}

 
 The \textit{Tucker} decomposition \cite{tucker1963implications,tucker1966some} is another way to decompose a tensor. It is a kind of higher order principal component analysis \cite{kolda2009tensor}. It decomposes tensor as a core tensor and factor tensors as Fig.~\ref{fig:tucker}. 
 
\noindent Let $X \in \mathbb{R}^{I \times J \times K}$, $A \in \mathbb{R}^{I \times P}, B \in \mathbb{R}^{J \times Q} , C \in \mathbb{R}^{I \times R}  \text{and} \ G \in \mathbb{R}^{P \times Q \times R} $   for $i = 1,\dots,I, j = 1,\dots,J,k = 1,\dots,K$. Here the size of the core tensor $P,Q$ and $R$ are the number of the components and it is hyperparameter. When the size of the core tensor is fixed, the size of the factor tensor is decided according to the size of the core tensor. Similar to the number of the components of \textit{CP} decomposition, we call deciding the size of the core tensor as a choosing the dimension. Then $x_{ijk}$ is approximated as follows. 

 \begin{equation}
      x_{ijk} \approx \sum_{p = 1}^{P}\sum_{q = 1}^{Q}\sum_{r = 1}^{R} g_{pqr}a_{ip}b_{jq}c_{kr}
 \end{equation}


\section{Method}
To mitigate the effect of the adversarial attacks, our insight stems from the tensor decomposition. 
In this section, all the paragraphs that describe revolve around how we can apply this magic, i.e., tensor decomposition,  as a defense mechanism against the adversarial attacks.

\subsection{Tensor decomposition as a preprocessing}
\begin{figure}[H]
    \centering
  \begin{subfigure}{0.24\columnwidth}
  \includegraphics[width=\textwidth]{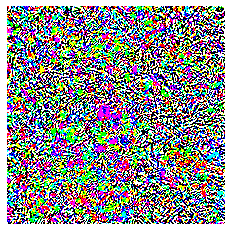}
  \caption{FGSM noise}
  \end{subfigure}
  \hfill
  \begin{subfigure}{0.24\columnwidth}
  \includegraphics[width=\textwidth]{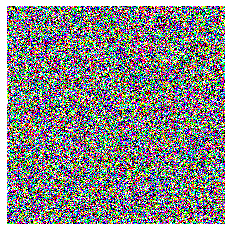}
  \caption{Gaussian noise} 
  \end{subfigure} 
  \begin{subfigure}{0.24\columnwidth} 
  \includegraphics[width=\textwidth]{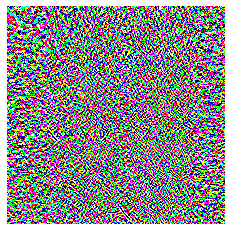}
  \caption{\textit{Tucker} noise} 
  \end{subfigure}  
  \hfill 
  \begin{subfigure}{0.24\columnwidth} 
  \includegraphics[width=\textwidth]{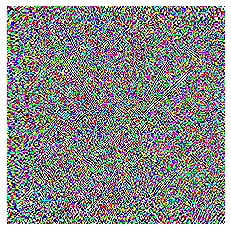}
  \caption{\textit{Cp} noise} 
  \end{subfigure}
  \caption{Visualization of example noises. (a) is the adversarial noise crafted by FGSM. (b) is the gaussian noise, (c),(d) are reconstructed images by using \textit{Tucker}, \textit{CP} decomposition respectively}
  \label{fig:noise}
  \end{figure}

 As you can see in Fig. \ref{fig:advs}, adversarial examples are too sophisticated to be recognized by human senses, including well-generalized deep learning models.
To prevent the potential threat, our model simply uses the reconstructed image from the tensor decomposition as an input.
We conjecture that the effect of the adversarial perturbations could be reduced by approximating the tensors toward the low dimension.
    To cast light on our hypothesis, we conduct brief experiments based on the visual sense.
  Fig. \ref{fig:noise} represents the examples of the noise. 
  Intuitively, adversarial noise crafted by the FGSM seems to distinguishable from others, while the rest of them are relatively similar to each other. 
  In other words, the tensor decomposition can transform adversarial noise into random noise, e.g., gaussian noise. 
We can say that the tensor decomposition has an ability to purify the adversarial noise in this regard.
Even though such random noise might also degrade the performance, it would be no worse than original adversarial noise considering they are crafted by adversarial intend.
   \vspace{-0.5cm}
   \begin{figure}[H]
    \centering
  \begin{subfigure}{\columnwidth}
  \includegraphics[width=\textwidth]{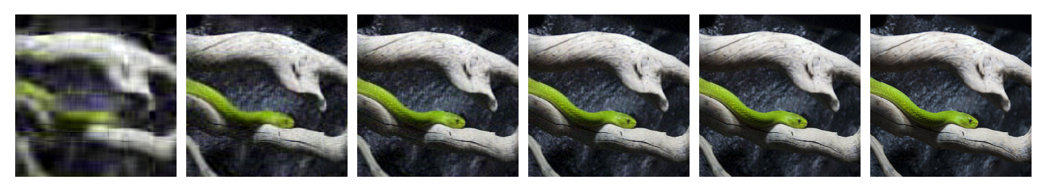}
  \caption{Reconstruction from \textit{CP} decomposition}
  \end{subfigure}
  \hfill
  \begin{subfigure}{\columnwidth}
  \includegraphics[width=\textwidth]{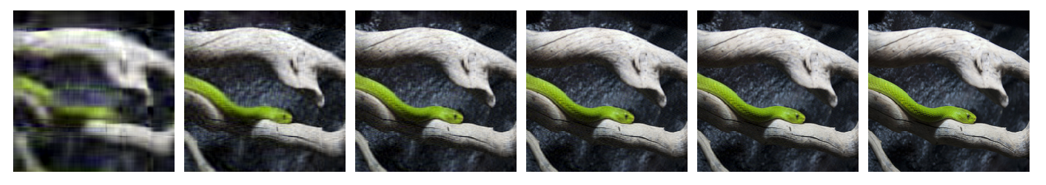}
  \caption{Reconstruction from \textit{Tucker} decomposition} 
  \end{subfigure} 
  \caption{We show a change of reconstructed images using \textit{CP} and \textit{Tucker} decomposition by varying the value of the dimension (The value is getting higher order by left $\rightarrow$ right)}
  \label{fig:gradation}
  \end{figure}
   Based on this light, we utilize \textit{CP} and \textit{Tucker} decomposition methods to verify how effective these methods actually are under various attack strategies.
Before the main experiments, both methods require to set the dimension, e.g., $r$ for the \textit{CP} and $P,Q$, $R$ for the \textit{Tucker}. 
As the dimension of the tensor increases, the quality of the reconstructed image gets better as shown in Fig. \ref{fig:gradation}.
The high quality of the reconstructed image is not always better, however, so the dimension needs to be decided in a heuristic manner.
We thus studied the ablation study to find out the effective hyperparameters under \textit{CP} and \textit{Tucker} and consider two kinds of factors to decide the hyperparameters, accuracy and time complexity.  
We randomly sampled 1,000 images from CIFAR-10, and then generate the adversarial images by applying FGSM, BIM, DeepFool, and C\&W, respectively.
As follow, those images are reconstructed by \textit{CP} and \textit{Tucker} decomposition.
Finally, we measured the accuracy and time complexity using the reconstructed images.
   \begin{figure}[H]
    \centering
  \begin{subfigure}{0.48\columnwidth}
  \includegraphics[width=\textwidth]{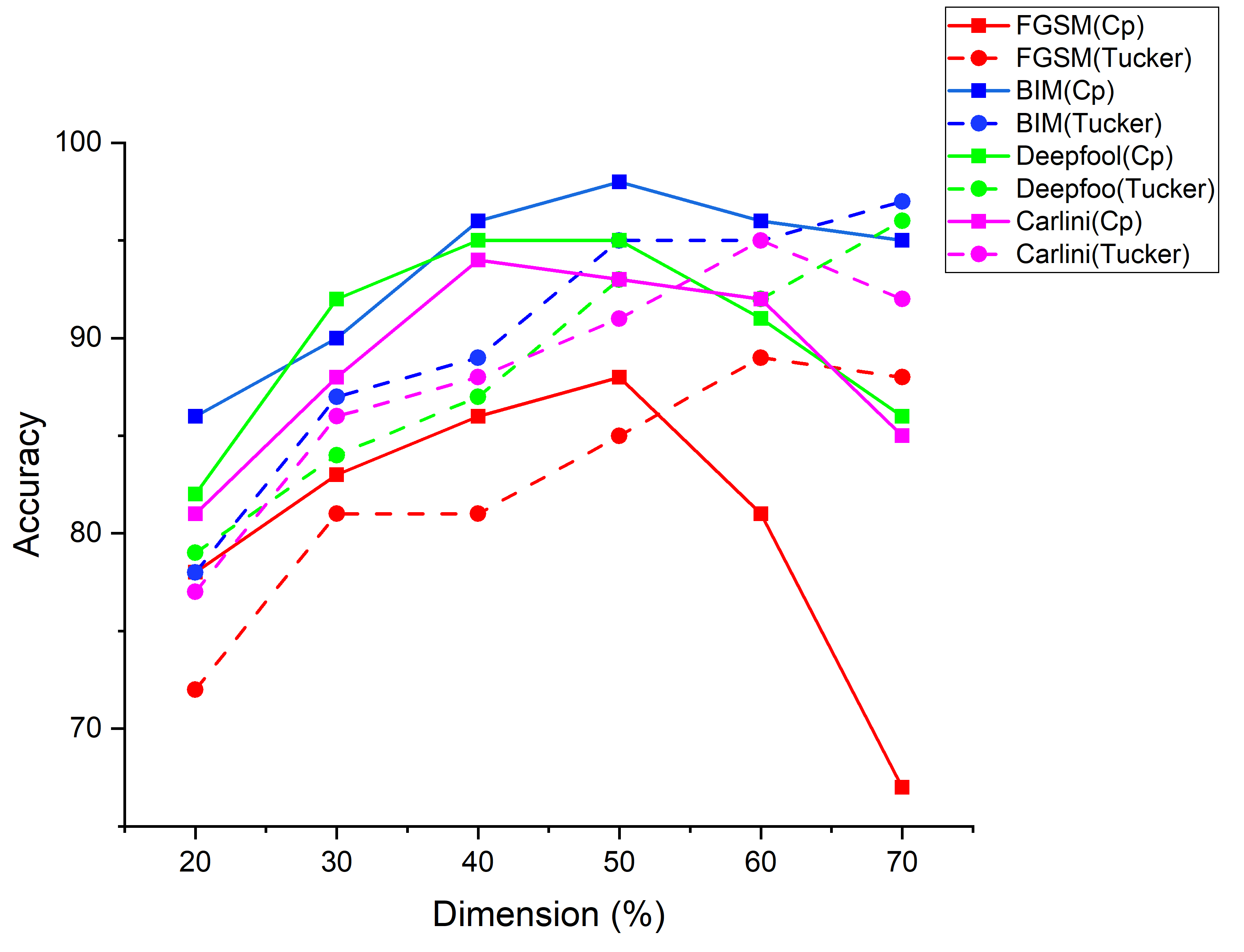}
  \caption{Top-1 accuracy}
  \end{subfigure}
  \begin{subfigure}{0.48\columnwidth}
  \includegraphics[width=\textwidth]{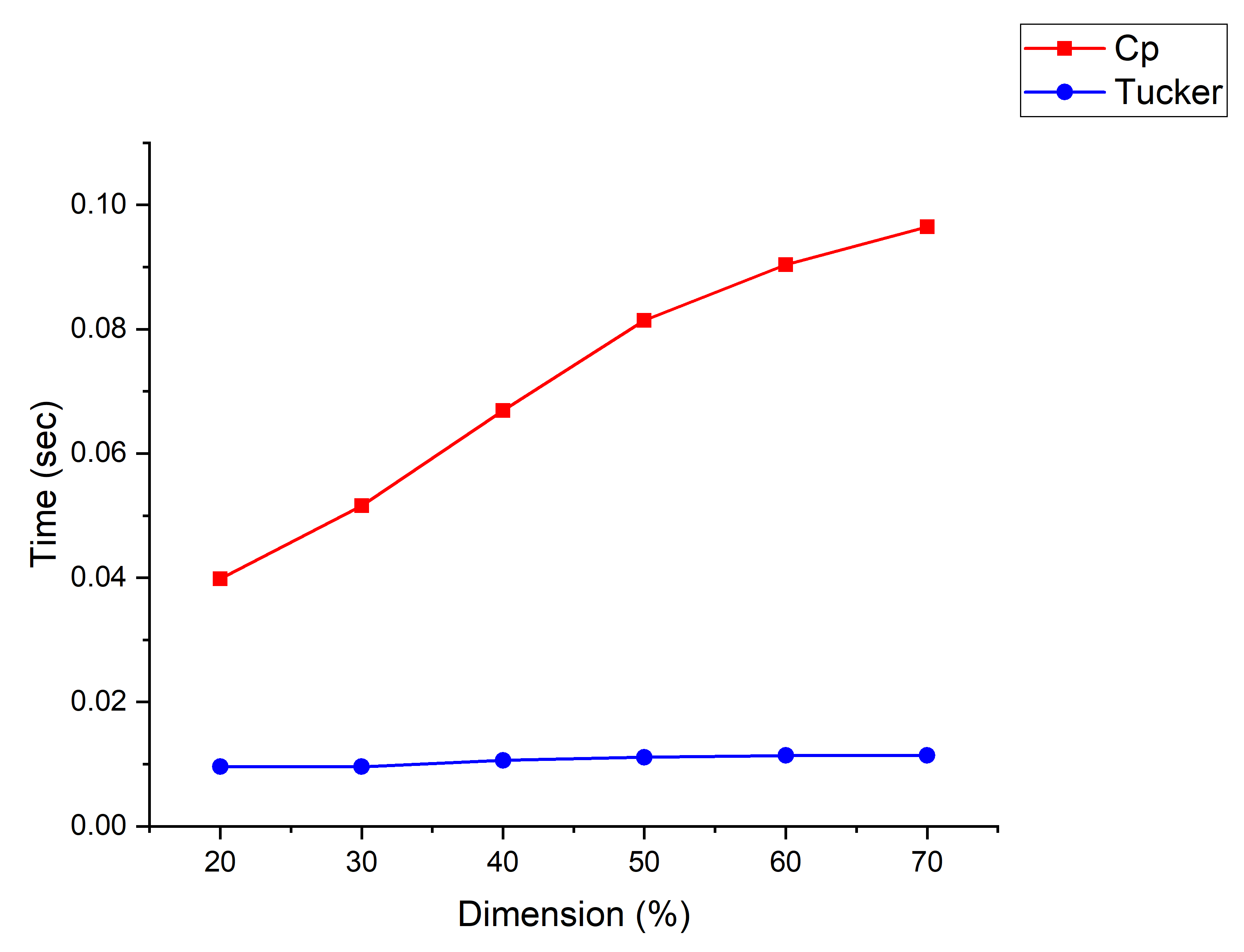}
  \caption{Processing time for single image} 
  \end{subfigure} 
  \caption{ (a) Accuracy with respect to dimension (b) Mean time for processing single image with respect to dimension}
  \label{fig:graph}
  \end{figure}  
The result summarized in Fig. \ref{fig:graph}. While accuracy increased as dimensions increased by the middle, accuracy tends to decrease gradually. And the processing time increases as dimension increases. So we decide to use 40 \% of the original dimension. For instance, the size of the image is 32 by 32 in the CIFAR-10 dataset.  In the case of \textit{CP} decomposition, the size of the three tensors in Fig. \ref{fig:cp} will be 32,32 and 3. So we choose rank $r = 8$. We use a similar argument when choosing the size of the core tensor of \textit{Tucker} decomposition. In particular, we don't compress the channel dimension $3$ in \textit{Tucker} decomposition since we don't want to lose color information. Thus, the size of the core tensor of the \textit{Tucker} decomposition is in the form of height, width and $3$. We use an open-source library \cite{kossaifi2019tensorly} for each decomposition method.

\subsection{Denoise Autoencoder as a supplement}

We should consider one more thing before putting the input value into the classifier.
 When it comes to reconstructing the images, we should consider the loss of information as it might affect the classification result.
 If the input is clean images, it would work even worse.
 To diminish the adverse effect from that point, we add denoise autoencoder into the procedure.
 Our method is based on a coarse-to-fine approach. 
 Through the reconstructed inputs from decomposed tensors, we remove the coarse-grained adversarial features. 
 We expect that some fine-grained features that might be lost by the coarse-grained approach—which is more likely to occur in high-dimension, could be compensated pass through the denoise autoencoder.
 Equipped with this approach, we set up the denoise autoencoder architecture as follows. The numerical value in Table \ref{table:autoencoder} stands for input channel and output channel respectively. And the filter size is $3 \times 3$.

 
\setlength{\tabcolsep}{6pt}
\begin{table}
\begin{center}
\caption{Architecture of autoencoder trained with each dataset}
\label{table:autoencoder}
\begin{tabular}{lclclclclclcl}
\hline\noalign{\smallskip}
MNIST & & & CIFAR & &\\
\noalign{\smallskip}
\hline
\noalign{\smallskip}
Conv,ELU & 1 & 2 & Conv,ELU & 3 & 6\\
Conv,ELU & 2 & 4 & Conv,ELU & 6 & 12\\
Conv,ELU & 4 & 8 & Conv,ELU & 12 & 24\\
Conv,ELU & 8 & 16 & Conv,ELU & 24 & 48\\
Conv,ELU & 16 & 32 & Conv,ELU & 48 & 96\\
Conv,ELU & 32 & 16 & Conv,ELU & 96 & 48\\
Conv,ELU & 16 & 8 & Conv,ELU & 48 & 24\\
Conv,ELU & 8 & 4 & Conv,ELU & 24 & 12\\
Conv,ELU & 4 & 2 & Conv,ELU & 12 & 6\\
Conv,ELU & 2 & 1 & Conv,ELU & 6 & 3\\
\hline
\end{tabular}
\end{center}
\end{table}
\setlength{\tabcolsep}{1.4pt}

We utilize the CIFAR-10 images at the RGB scale and MNIST image at the grayscale.
We set the learning rate to $1e-4$ and used Adam \cite{kingma2014adam} as an optimizer. 
And we use mean square error (MSE) for loss function. 
For both models, we train autoencoder for 10 epochs. We henceforth denote autoencoder as \textit{AE}.
 
\subsection{Overall architecture}

\begin{figure}[H]
\centering
\includegraphics[width=11cm,height = 6cm]{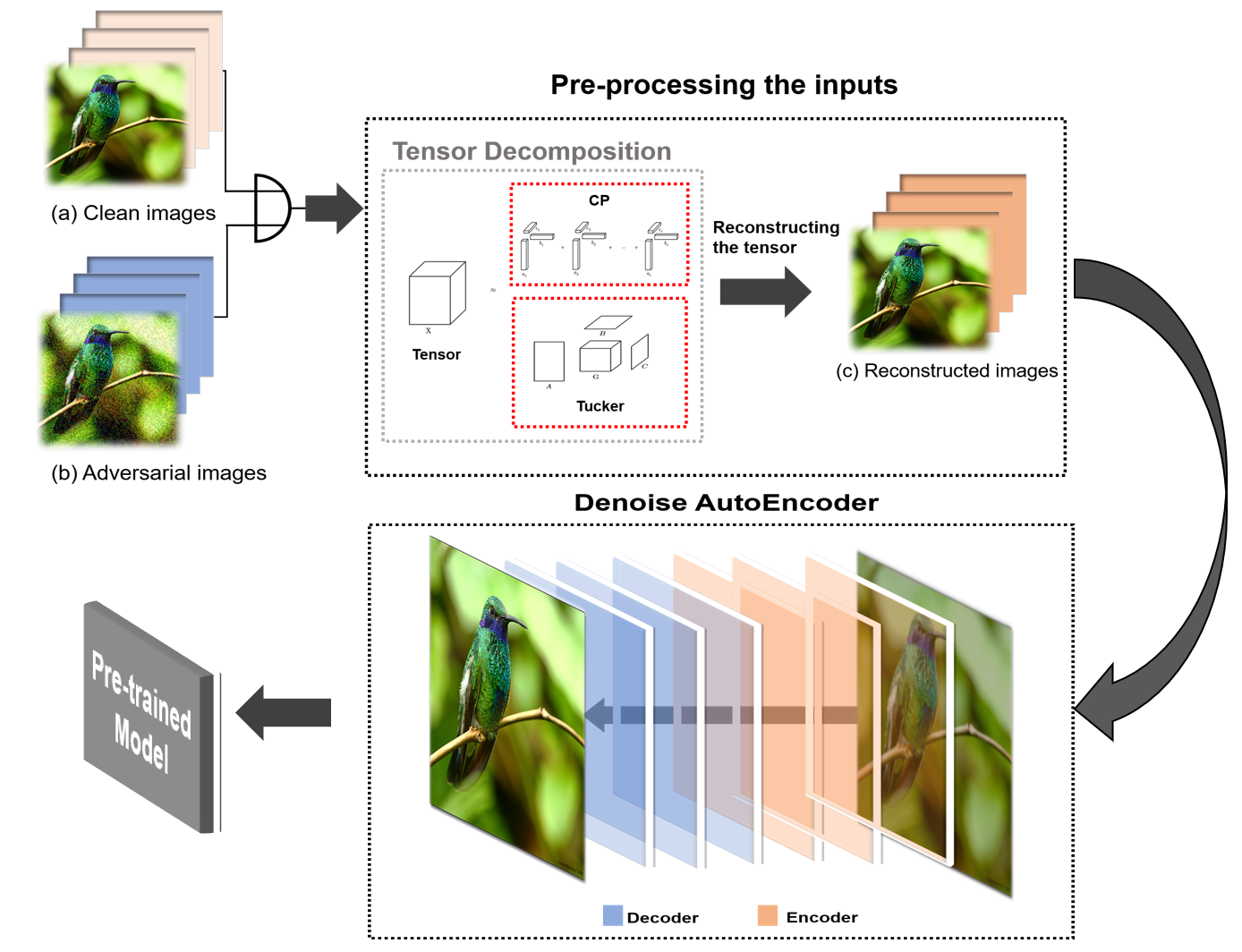}
\caption{Overall flow of our proposed method}
\label{fig:architecture}
\end{figure}

We describe the overall architecture in detail.
Fig.~\ref{fig:architecture} represents the overall flow of our proposed method. 
First, we approximate the input image via the tensor decomposition method. 
The inputs could be adversarial images or clean images. 
Our method does not spend time deciding whether the input is adversarial or not.
That's the reason why our model does not require a well-generalized model.
In other words, whatever the input is, our model splits the input into several tensors based on \textit{CP} or \textit{Tucker} decomposition, and then reconstruct them.
As follows, the reconstructed images are passed through the denoise autoencoder, which might compensate for losing the information that may in charge of an important role in that image.
Note that our method does not have a model dependency, so it can be applied in conjunction with every classifier. 

\section{Experiment}
\subsection{Dataset}

 For the MNIST, CIFAR-10 data, we test on the full test data, which are composed of  10,000 images on MNIST and 50,000 images on CIFAR-10. For ImageNet, we select randomly 1,000 images as similar setting \cite{jia2019comdefend,kurakin2018adversarial}. Since our proposed method decomposes the input whichever clean image or adversarial image, we tested on both clean images and adversarial images.

\subsection{Evaluation} 
We measure the top-1 accuracy on clean images and adversarial images on each dataset. For MNIST, we use a simple model consists of two convolutional layers. For CIFAR-10 and ImageNet, we basically use pre-trained Resnet101 \cite{he2016deep}. In particular, for CIFAR-10, we finetune pre-trained Resnet101 for 10 classes.  We use FGSM, BIM, DeepFool and C\&W attack methods. For the distance metric, a related research area mainly uses $L_{\infty}$ and $L_2$ norm \cite{carrara2018adversarial,jia2019comdefend}. In detail, we use $L_{\infty}$ for FGSM, BIM, and DeepFool attack. And for the C\&W attack, we use a $L_2$ metric. We try to find small perturbation when applying the adversarial attack since the noise is visible when the perturbation is not small enough. We generate adversarial images by using open source library Foolbox \cite{rauber2017foolbox}. In detail, we try 100 epsilons from 0 to 1 for FGSM. For BIM, we set 5 as a number of iteration. And for DeepFool, we set 50 as a maximum number of steps and set 50 as a maximum iteration number of C\&W.  And we measure the pre-processing time for calculating the additional time consuming for the proposed method. Also, we compare our results to other state-of-the-art defense models. For a fair comparison, we compare the ratio between the accuracy of the clean image and the adversarial image since the accuracy of a clean image is a little bit different depending on the setting.
 
\subsection{Results}

We achieve remarkably high accuracy against adversarial attacks. In most cases, the \textit{CP} is better than \textit{Tucker} decomposition method. In some case of ImageNet dataset, \textit{Tucker} decomposition method is better than \textit{CP}. For instance, when attack with FGSM and C\&W method, the result was the best by using \textit{Tucker} decomposition.  And in the case of clean images, the accuracy reduction was about 1\% on all datasets. It means that we do not harm the original model in a normal case which is the input image is clean. The autoencoder is highly effective on MNIST dataset. Although the autoencoder does not have much effect on clean images, it improves the performance of various adversarial attacks on MNIST dataset. In addition to MNIST dataset, there have been small performance improvements for other datasets by using the denoise autoencoder. The numerical results are summarized in Table
 \ref{table:MNIST}, \ref{table:CIFAR} and \ref{table:Imagenet}. 
 
 \setlength{\tabcolsep}{6pt}
\begin{table}[H]
\begin{center}
\caption{Top-1 accuracy of each method on MNIST dataset}
\label{table:MNIST}
\begin{tabular}{lllllll}
\hline\noalign{\smallskip}
Model & Method & Clean & FGSM & BIM & DeepFool & C\&W\\
\noalign{\smallskip}
\hline
\noalign{\smallskip}
\multirow{5}{*}{2-Conv Net} & Original  & 99.06 & 0.00 & 0.21 & 0.00 & 0.00 \\
 & CP  & 99.01& 71.62 & 95.42& 81.68 & 88.29\\
 & CP+AE & 98.64 & 96.15 & 98.88 & 98.83 & 98.6\\
 & Tucker & 99.06 & 71.95 & 91.81 & 79.32 & 84.52\\
 & Tucker+AE & 98.06 & 95.26 & 98.65 & 98.54 & 98.05\\
\hline
\end{tabular}
\end{center}
\end{table}
\vspace*{-1.5cm}
\setlength{\tabcolsep}{6pt}
\begin{table}[H]
\begin{center}
\caption{Top-1 accuracy of each method on CIFAR-10 dataset}
\label{table:CIFAR}
\begin{tabular}{lllllll}
\hline\noalign{\smallskip}
Model & Method & Clean & FGSM & BIM & DeepFool & C\&W\\
\noalign{\smallskip}
\hline
\noalign{\smallskip}
\multirow{5}{*}{Resnet-101}& Original & 98.87 & 0.0 & 0.0 & 0.0 & 0.0 \\
  & CP & 97.96 & 92.52 & 92.49 &  94.67 & 93.63\\
  & CP+AE & 98.11 & 92.88 & 93.21 & 94.93 & 94\\
  & Tucker & 95.99 & 87.74 & 87.68 & 90.73 & 90.57\\
  & Tucker+AE & 96.00 & 88.44 & 88.56 & 91.28 & 91.2 \\
\hline
\end{tabular}
\end{center}
\end{table}

\vspace*{-1.5cm}

\setlength{\tabcolsep}{6pt}
\begin{table}[H]
\begin{center}
\caption{Top-1 accuracy of each method on ImageNet dataset}
\label{table:Imagenet}
\begin{tabular}{lllllll}
\hline\noalign{\smallskip}
Model & Method & Clean & FGSM & BIM & DeepFool & C\&W\\
\noalign{\smallskip}
\hline
\noalign{\smallskip}
\multirow{5}{*}{Resnet-101}& Original & 77.6 & 0.0 & 0.0 & 0.0 & 0.0\\
  & CP & 76.3 & 61.2 & 75.4 & 75.7 & 61.2\\
  & CP+AE & 74.8& 62.9 & 75.00 & 75.5 & 62.1\\
  & Tucker & 76.4 & 65.00 & 73.4 & 73.8 & 65.3 \\
  & Tucker+AE & 74.7 &  65.6 & 73.7 & 74.7 & 66\\
\hline
\end{tabular}
\end{center}
\end{table}
\setlength{\tabcolsep}{1.4pt}

Even the DeepFool and C\&W attacks are more accurate and powerful attack compared to the FGSM and BIM attacks, the accuracy after decomposition is higher than the case of FGSM and BIM attacks. 

\subsection{Comparison with other defense methods}
\begin{figure}[H]
    \centering
  \begin{subfigure}{0.48\columnwidth}
  \includegraphics[width=\textwidth]{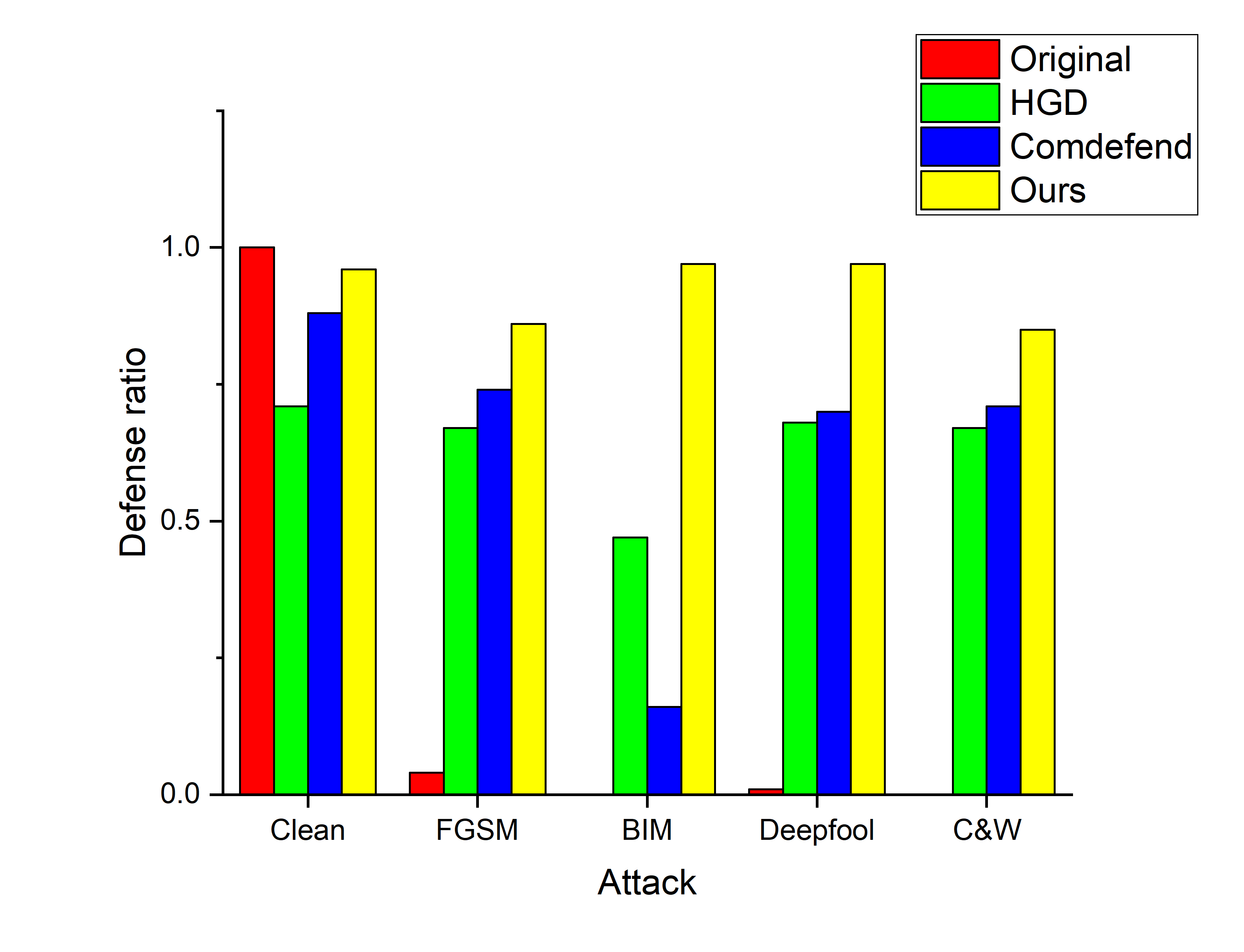}
  \caption{Resnet101 on ImageNet}
  \end{subfigure}
  \begin{subfigure}{0.48\columnwidth}
  \includegraphics[width=\textwidth]{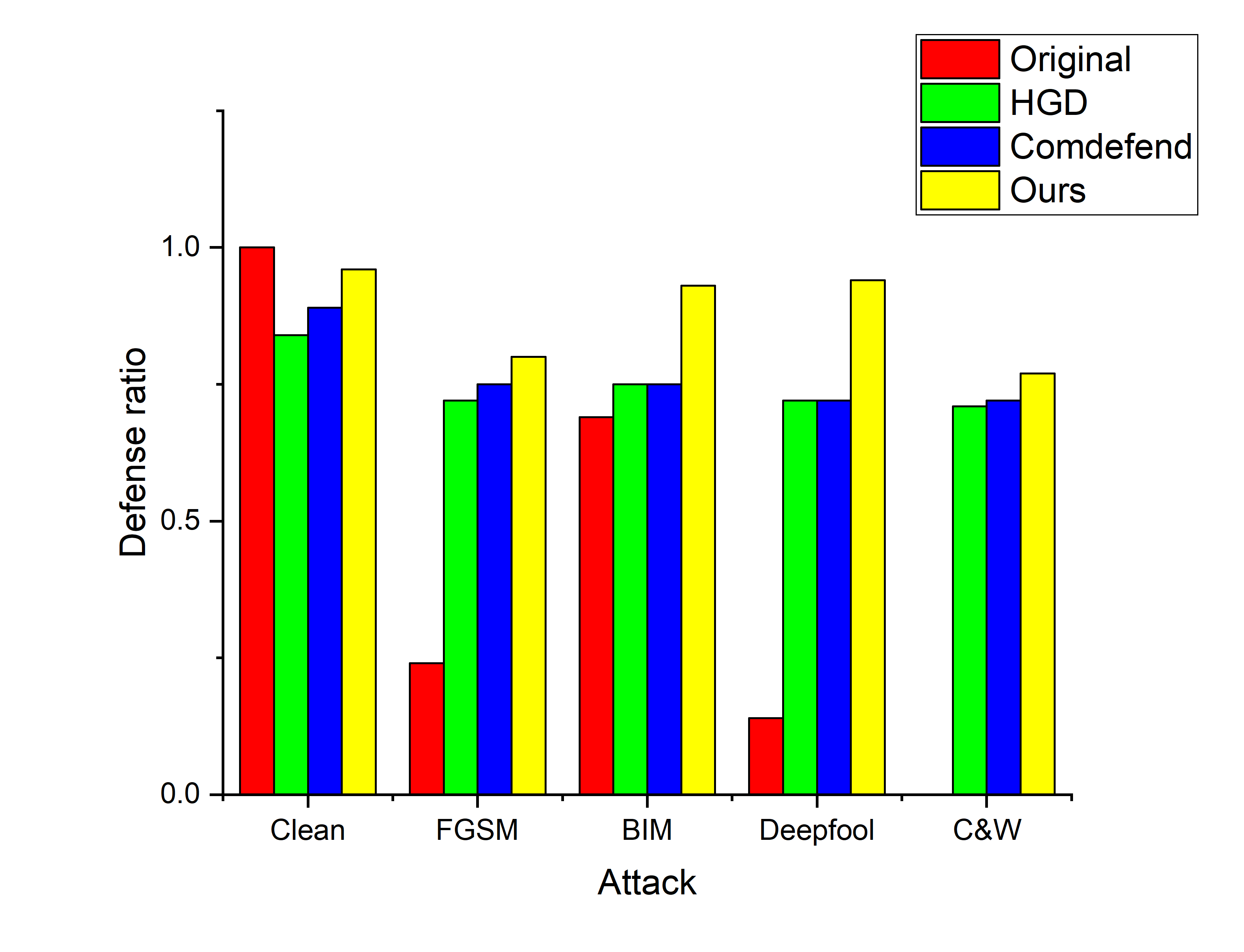}
  \caption{IncV3 on ImageNet} 
  \end{subfigure} 
  \caption{ (a) Defense ratio of each attack method on Resnet101  (b) Defense ratio of each attack method on IncV3. Note that our method outperforms other defense methods}
  \label{fig:bar}
  \end{figure}

 We measure the ratio of accuracy on clean images and adversarial images generated by FGSM, BIM, DeepFool and C\&W attack for a fair comparison. Here the $L_{\infty}$ is restricted to $8$ in $255$ scale. The defense ratio is defined as follows.
 \begin{equation}
     \textit{Defense ratio} = \frac{\textit{Top-1 accuracy on adversarial images}}{\textit{Top-1 accuracy on clean images}} 
 \end{equation}
 We compare the performance of recent defense methods, HGD \cite{Liao_2018_CVPR} and Comdefend \cite{jia2019comdefend}. Fig. \ref{fig:bar} shows the results. We select Resnet101 \cite{he2016deep} and Inception V3 (IncV3) \cite{szegedy2016rethinking} as base model. And we tested on 1,000 images from the ImageNet data. Our methods outperform in all cases compared to two recent defense methods. This result verifies that our method is effective. Moreover, our method does not depend on attack methods and the target model classifier, thus it can be easily combined with every model.

\subsection{Time analysis}
 We measure the preprocessing time of each method. We pick randomly 1,000 images in MNIST,CIFAR-10, and ImageNet. And we calculate the average processing time per image. In most case, the \textit{CP} decomposition requires more time compared to \textit{Tucker} decomposition. In the case of MNIST and CIFAR-10, the time required to reconstruction is similar in both cases. However, In the case of ImageNet, the \textit{CP} decomposition takes about 10 times more than the \textit{Tucker} method. Table \ref{table:time} summarizes preprocessing time of each dataset on each method.

\setlength{\tabcolsep}{7pt}
\begin{table}[H]
\begin{center}
\caption{Preprocessing time of the each dataset}
\label{table:time}
\begin{tabular}{*5c}
\hline\noalign{\smallskip}
dataset&CP&CP+AE&Tucker&Tucker + AE\\
\noalign{\smallskip}
\hline
\noalign{\smallskip}
  MNIST & 0.005 & 0.006 & 0.003 & 0.004\\
  CIFAR-10 & 0.1052 & 0.1161 & 0.01268 & 0.015 \\
  ImageNet & 1.07 & 1.18 & 0.1566 & 0.17 \\
\hline
\end{tabular}
\end{center}
\end{table}

\subsection{White box scenario}
In the white box scenario, we should assume the adversary knows full defense mechanism according to \cite{carlini2017adversarial}. In our method, note that the input image is always decomposed and reconstructed, and the decomposed components are always started from the random tensor. In detail, the component tensors of each decomposition method initialized to random tensor and then trained to approximate the original tensor. Thus, the input is always random tensor and the original image is a label itself like unsupervised learning. Therefore, there are no fixed weights, so the adversary can not generate adversarial examples concerning the tensor decomposition method. Hence, our propose method is robust on the white box attack scenario.

\section{Conclusions}

In this work, we verify the tensor decomposition is a simple and powerful method for purifying the adversarial perturbation. When we combine denoise autoencoder with the tensor decomposition method, the proposed method achieves higher accuracy against adversarial attacks. We experiment with our method against various adversarial attacks such as DeepFool and C\&W attacks and discuss why this method is robust in the white box scenario. 
 
Our intuition applying tensor decomposition into the adversarial attack is as follows. Since the adversarial perturbation is so small that it is hard to catch a difference, such a small perturbation would be removed by approximating the image tensor using low dimensional tensors. Since there is no straightforward algorithm to choose the dimension of the component tensors of the \textit{CP} and \textit{Tucker} decomposition, finding the best dimension remains for future work. Also, establishing a theoretical base why tensor decomposition is robust against adversarial attack is left to our future work.

\clearpage
%
%
\bibliographystyle{splncs04}
\bibliography{egbib}

\end{document}